\documentclass[10pt,twocolumn,letterpaper]{article}

\usepackage[pagenumbers]{wacv} 

\usepackage{graphicx}
\usepackage{amsmath}
\usepackage{amssymb}
\usepackage{booktabs}

\usepackage[pagebackref,breaklinks,colorlinks]{hyperref}

\usepackage{tabularx}
\newcolumntype{Y}{>{\centering\arraybackslash}X}

\usepackage[table]{xcolor}
\definecolor{mygray}{RGB}{169, 169, 169}
\definecolor{mypink}{RGB}{245, 171, 186}
\definecolor{myblue}{RGB}{91, 208, 250}
\definecolor{mydarkblue}{RGB}{0, 153, 204}

\usepackage[capitalize]{cleveref}
\usepackage{relsize}
\crefname{section}{Sec.}{Secs.}
\Crefname{section}{Section}{Sections}
\Crefname{table}{Table}{Tables}
\crefname{table}{Tab.}{Tabs.}

\def\wacvPaperID{1110} 
\def\confName{WACV}
\def\confYear{2025}

\begin{document}

\title{Enhancing Scene Graph Generation with Hierarchical \\ Relationships and Commonsense Knowledge}

\author{Bowen Jiang, Zhijun Zhuang, Shreyas S. Shivakumar, Camillo J. Taylor\\
GRASP Laboratory, University of Pennsylvania\\
Philadelphia, PA, 19104, USA\\
{\tt\small \{bwjiang, zhijunz, sshreyas, cjtaylor\}@seas.upenn.edu}}
\maketitle

\begin{abstract}
This work introduces an enhanced approach to generating scene graphs by incorporating both a relationship hierarchy and commonsense knowledge. Specifically, we begin by proposing a hierarchical relation head that exploits an informative hierarchical structure. It jointly predicts the relation super-category between object pairs in an image, along with detailed relations under each super-category. Following this, we implement a robust commonsense validation pipeline that harnesses foundation models to critique the results from the scene graph prediction system, removing nonsensical predicates even with a small language-only model. Extensive experiments on Visual Genome and OpenImage V6 datasets demonstrate that the proposed modules can be seamlessly integrated as plug-and-play enhancements to existing scene graph generation algorithms. The results show significant improvements with an extensive set of reasonable predictions beyond dataset annotations.
Codes are available at \href{https://github.com/bowen-upenn/scene_graph_commonsense}{https://github.com/bowen-upenn/scene\_graph\_commonsense}\footnote{This work has been accepted at the 2025 IEEE/CVF Winter Conference on Applications of Computer Vision (WACV).}.

\end{abstract}    
\section{Introduction}
This work presents simple yet effective approaches in the field of scene graph generation~\cite{johnson2015image, lu2016visual, jiang2023hierarchical, zhu2022scene, chang2021comprehensive}. Scene graph generation, a complex problem that deduces both objects in an image and their pairwise relationships, moves beyond object detection or segmentation methods~\cite{redmon2016you, ren2015faster, carion2020end, zhang2024deepgi, li-19-segmentation} which isolate individual object instances. Instead, it represents the entire image as a graph, where each object instance forms a node and the relationships between nodes form directed edges.

Existing literature has addressed the nuanced relationships in visual scenes by designing sophisticated architectures~\cite{dhingra2021bgt, lin2020gps, xu2021joint, cong2022reltr, ren2020scene, chen2019knowledge}. This work shows how the performance of these methods can be enhanced by exploiting a natural hierarchy among the relationship categories.
Adopting the definitions in Neural Motifs~\cite{zellers2018neural} to divide predominant relationships in scene graphs into \textit{geometric}, \textit{possessive}, and \textit{semantic} super-categories, we show how these categories can be explicitly utilized in a network. We also explore automatic clustering in the token embedding space without human involvement. As a result, our proposed hierarchical classification scheme aims to jointly predict the probabilities of relation super-categories and the conditional probabilities of relations within each super-category. 

\begin{figure}[t]
  \centering
  \includegraphics[width=1\linewidth]{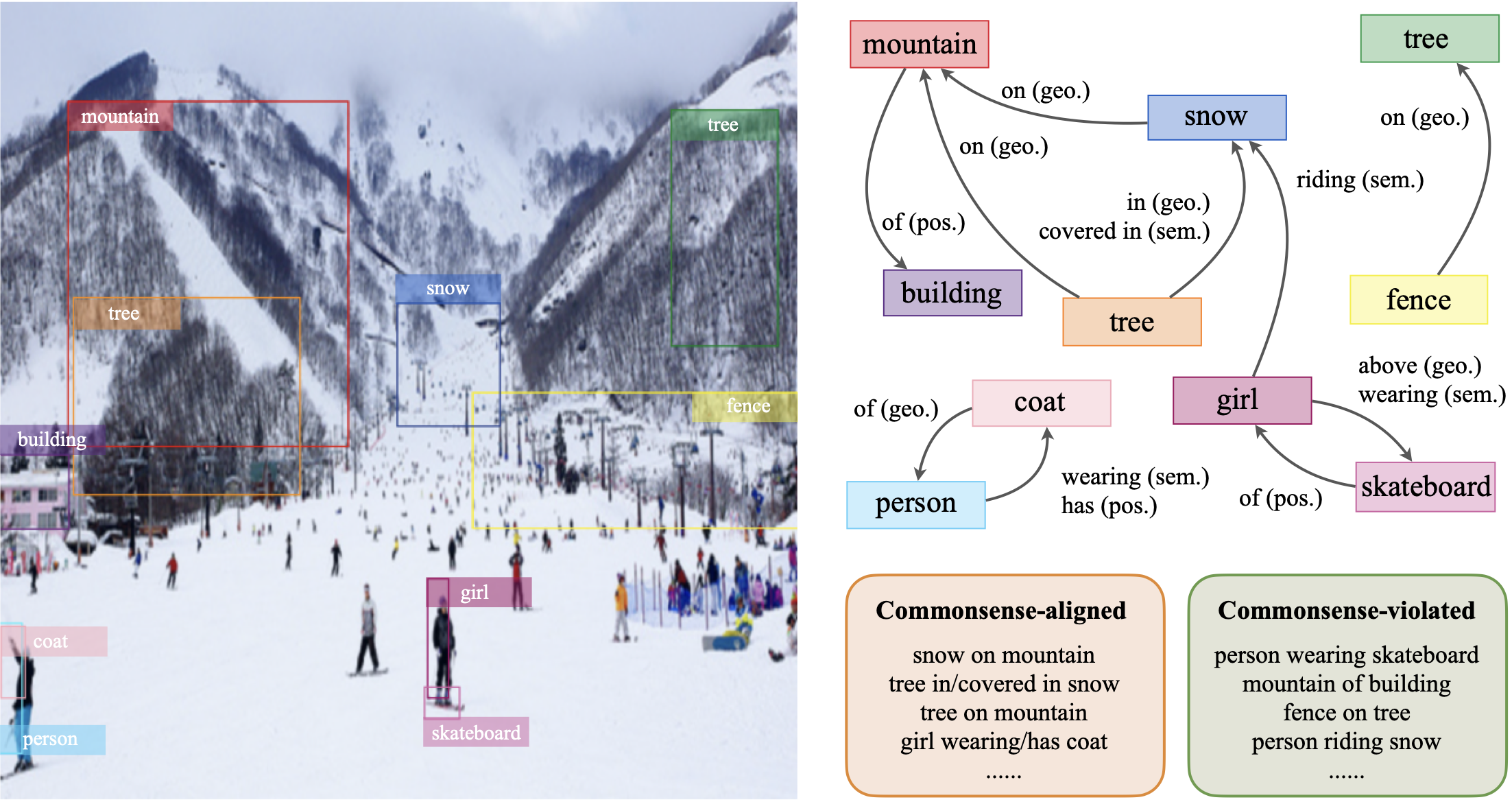}
  \caption{Example scene graph. Each edge represents a predicted relationship under a geometric, possessive, or semantic super-category, and it can be either commonsense-aligned or violated.}
  \vspace{-3mm}
  \label{fig:top}
  \setcounter{figure}{1}  
\end{figure}

Although an advanced scene graph generation model may achieve good performance indicated by its high recall scores~\cite{lu2016visual, tang2019learning}, it may produce a wide range of unreasonable relationships that are unlikely to occur in the real world, such as \textit{"bunny jumping plate"}, even with a high confidence. Recent advances in large language models (LLMs)~\cite{brown2020language, achiam2023gpt, touvron2023llama, team2023gemini, anthropic2024claude, jiang2023mistral, anil2023palm} and vision-language models (VLMs)~\cite{OpenAI2023GPT4V, liu2024visual, liu2023improved, jiang2024multi, reid2024gemini, wang2023cogvlm, lu2024chameleon} now enables machines to perform commonsense reasoning~\cite{wei2022chain, bubeck2023sparks, zhao2024large}. Therefore, we incorporate LLMs or VLMs into the system to critique the output of a scene graph generation model, removing predicates that are not accord with common sense intuitions.

In this study, we present the HIErarchical Relation head and COMmonsense validation pipeline, \textbf{HIERCOM}. Comprehensive experimental results show that these straightforward, plug-and-play modules can substantially enhance the performance of existing scene graph generation models, often by a large margin. Improvements are observed across Recall@$k$~\cite{lu2016visual}, mean Recall@$k$~\cite{tang2019learning}, and zero-shot evaluation metrics. These two innovations enable even a simple baseline model to produce reliable, commonsense-aligned outcomes, and elevate state-of-the-art (SOTA) methods to new levels of performance. Furthermore, our findings reveal that language models, regardless of their scale or whether they are augmented with vision capabilities, perform robustly in commonsense validation tasks. This consistency in strength makes our algorithms more accessible, allowing the community to deploy HIERCOM with just a small-scale, language-only model on local devices for efficient scene graph generation.

\section{Related Work}
\label{sec:related}

\paragraph{Scene graph generation with hierarchical information}
Neural Motifs~\cite{zellers2018neural}, as an early work, analyzes the Visual Genome~\cite{krishna2017visual} dataset and divide the $50$ most frequent relations into $3$ super-categories: \textit{geometric}, \textit{possessive}, and \textit{semantic}, as detailed in Figure~\ref{fig:histogram}. Unfortunately, it does not further utilize the super-categories it identifies. Our work follows their definitions and fully utilizes this hierarchical structure. Besides, HC-Net~\cite{ren2020scene} investigates hierarchical contexts, rather than the relations. \cite{neau2023fine} studies the fine-grained relations and explores different usage of each predicate in practice. GPS-Net~\cite{lin2020gps} focuses on understanding the relative priority of the graph nodes. CogTree~\cite{yu2020cogtree} and HML~\cite{deng2022hierarchical} focus on automatically building a tree structure from coarse to fine-grained levels, while relations in our work are clustered by their semantic meanings either manually~\cite{zellers2018neural} or automatically using token embeddings. 
\vspace{-1mm}


\paragraph{Scene graph generation with foundation models}
ELEGANT~\cite{zhao2023less} leverages LLMs to propose potential relation candidates based on common sense and validate them using BLIP-2~\cite{li2023blip}. \cite{zhang2024enhancing} collaborates multiple LLMs for enhanced scene understanding and LLM-AR~\cite{qu2024llms} utilizes LLMs as action recognizers. \cite{yao2021visual} uses CLIP~\cite{radford2021learning} to verify extracted triplets from knowledge bases, and RECODE~\cite{li2024zero} further aids CLIP with visual cues from LLMs. \cite{mitra2024compositional} prompts VLMs to generate scene graphs in JSON formats. VLM4SGG~\cite{kim2024llm4sgg} and other works \cite{zhong2021learning, li2022integrating, ye2021linguistic} allow language supervisions for weakly supervised generation, and \cite{li2024pixels, yu2023visually, he2022towards, zhang2023learning} further extend the scope to open vocabularies.
\vspace{-1mm}

\paragraph{Other scene graph generation methods}
Scene graph generation was initially proposed in \cite{johnson2015image, lu2016visual}. Many approaches tackle the problem from the perspective of graphical neural networks~\cite{scarselli2008graph, xu2017scene, yang2018graph, lin2020gps, chen2019knowledge, khademi2020deep, zareian2020weakly, ulger2023relational} or recurrent networks~\cite{zellers2018neural, xu2017scene, cho2014properties, dhingra2021bgt} to integrat global contexts. In recent years, \cite{khandelwal2022iterative} shifts the iterative message passing to transformers. EGTR~\cite{im2024egtr} extracts relations from self-attention layers of the DETR~\cite{carion2020end} decoder. BGT-Net~\cite{dhingra2021bgt} and RTN~\cite{koner2020relation} integrates two transformers for objects and edges, respectively. RelTR~\cite{cong2022reltr} and SSR-CNN~\cite{teng2022structured} further extends the notions to triplets. \cite{suhail2021energy} offers an energy-based framework.
\cite{xu2021joint} establishes a conditional random field to model the distribution of objects and relations. \cite{tang2020unbiased} aims to remove the bias from prediction via total direct effects. 

Furthermore, there are a series of works paying specific attention to the long-tailed distributions of relations~\cite{tang2020unbiased, li2022devil, yu2020cogtree, zhang2022fine, yu2023visually, li2021bipartite, goel2022not, li2022ppdl, deng2022hierarchical, neau2023fine, lorenz2023haystack}. Most works sacrifice R@$k$~\cite{lu2016visual} for mR@$k$~\cite{tang2019learning} scores. However, our work presents plug-and-play approaches that are \textit{model-agnostic}, so it can incorporate with those models - specifically designed for reducing imbalance - to improve both their R@$k$ and mR@$k$ scores simultaneously. We show experiments in Section~\ref{sec:long_tail} and the detailed histogram in Figure~\ref{fig:histogram}.

\section{Scene Graph Construction}
\label{sec:method}
A scene graph $G=\{V, E\}$ is a graphical representation of an image. The set of vertices $V$ consists of $n$ object instances, including their bounding boxes and labels. The set of edges $E$ consists of one or more relationships $\mathbf{r}$, if any. 

This section first introduces a standalone baseline model with a better evaluation flexibility. It then describe the proposed hierarchical relation head and commonsense validation pipeline, which are designed as plug-and-play modules that can continue enhancing existing SOTA scene graph generation methods to new levels of performance.

\begin{figure*}[t]
  \centering
  \includegraphics[width=0.99\linewidth]{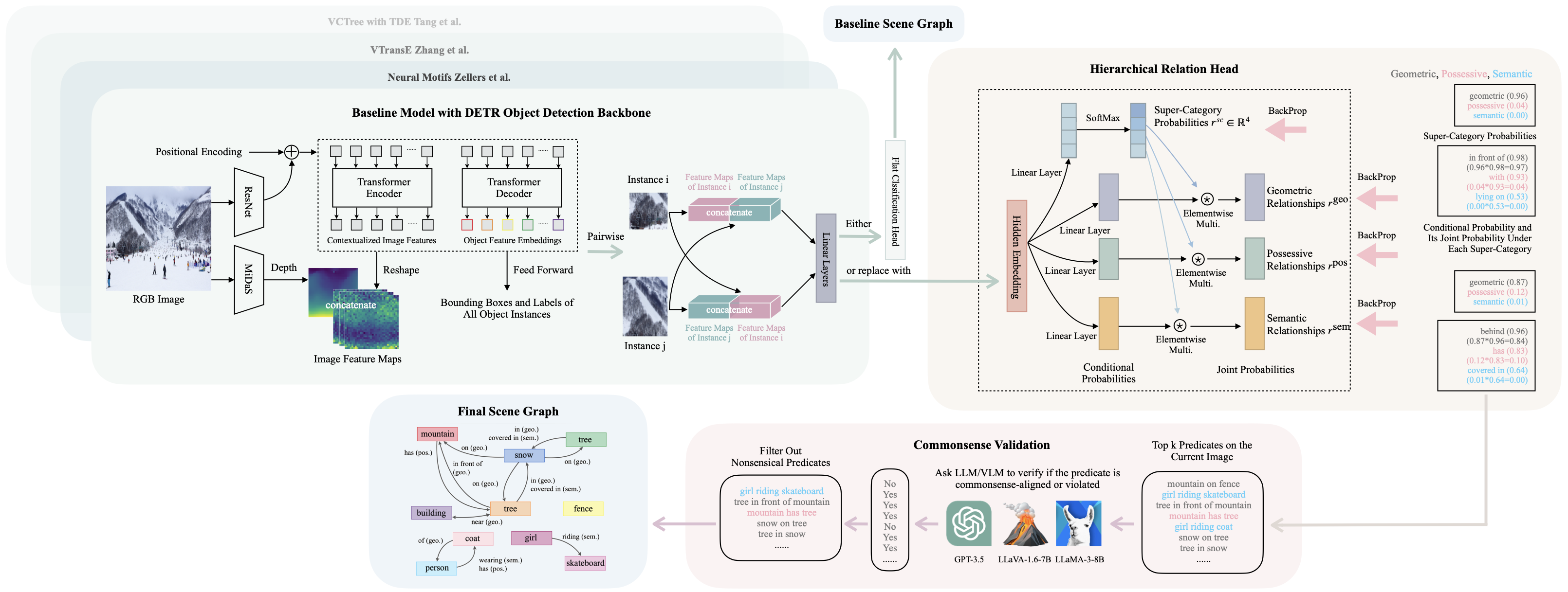}
  \caption{This diagram provides an overview of HIERCOM. The core components of HIERCOM are the hierarchical relation head and the commonsense validation pipeline, both of which are model-agnostic plug-and-play modules, suitable for integration with a variety of baseline scene graph generation models that have a conventional flat classification head. Specifically, the hierarchical relation head is designed to replace this flat layer, jointly estimating relation super-categories and more granular relations within each category. Additionally, the diagram depicts a baseline scene graph generation model: an RGB image and depth maps are the inputs, and it generates feature maps, object labels, and bounding boxes using DETR. Relations estimation between each pair of instances occurs in two separate passes to account for directional relationships, first assuming one instance as the subject and then the other. Subsequently, the commonsense validation pipeline leverages an LLM or VLM - which can have a small size - to filter out commonsense-violating predicates}
  \label{fig:flowchart}
  \vspace{-3mm}
\end{figure*}


\subsection{Baseline model}
We propose a simple yet strong baseline model that adapts the widely-used two-stage design~\cite{xu2017scene, yang2018graph, zellers2018neural, lin2020gps, dhingra2021bgt}. It starts with a Detection Transformer (DETR)~\cite{carion2020end} as the object detector with a ResNet-101 backbone~\cite{he2016deep}. Its transformer encoder~\cite{carion2020end, vaswani2017attention} can contextualize image features $\boldsymbol{I} \in \mathbb{R}^{h \times s \times t}$ with global information at an earlier stage of the scene graph generation pipeline. Here, $h$ is the hidden channels, while $s$ and $t$ denote the spatial dimensions. Its decoder outputs a set of object bounding boxes and labels in parallel. We also integrate MiDaS~\cite{ranftl2020towards} to estimate a depth map $\boldsymbol{D} \in \mathbb{R}^{s \times t}$ from each single image, which will be concatenated with $\boldsymbol{I}$ as ${\boldsymbol{I}^{\prime}} \in \mathbb{R}^{(h + 1)\times s \times t}$.

The model computes hidden embeddings for both the subject and object instances, separately, to overcome
possible overlaps between them and maintain their relative
spatial locations. These embeddings are obtained by dot-multiplying their respective bounding boxes $\boldsymbol{M}_{i}, \boldsymbol{M}_{j} \in \mathbb{R}^{s\times t}$ with ${\boldsymbol{I}^{\prime}}$, resulting in two feature tensors $\boldsymbol{I}_{i}^{\prime}, \boldsymbol{I}_{j}^{\prime} \in \mathbb{R}^{(h+1)\times s \times t}$. To address the directional nature of relationships, $\boldsymbol{I}_{i}^{\prime}$ and $\boldsymbol{I}_{j}^{\prime}$ are concatenated in both possible directions as $\boldsymbol{I}_{ij}^{\prime}$ and $\boldsymbol{I}_{ji}^{\prime} \in \mathbb{R}^{2\cdot (h+1)\times s \times t}$. The concatenated tensors are then fed separately into subsequent linear layers as $\boldsymbol{X}_{ij}$ and $\boldsymbol{X}_{ji} \in \mathbb{R}^{d}$, and the classification head will finally estimate the relations $\mathbf{r}_{ij}$ and $\mathbf{r}_{ji}$, as shown in Equation~\ref{eqn:flat}.
\vspace{-1mm}
\begin{equation}
	\boldsymbol{r}_{ij},\ \boldsymbol{r}_{ji} = \operatorname{SoftMax}\{\boldsymbol{X}_{ij}^{\top} \boldsymbol{W}\},\ \operatorname{SoftMax}\{\boldsymbol{X}_{ji}^{\top} \boldsymbol{W}\}\label{eqn:flat}
\end{equation}
\vspace{-1mm}
where $\boldsymbol{W}$ is a learnable parameter tensor of the linear layer.

\subsection{Hierarchical relation head}
Inspired by the Bayes' rule, the hierarchical relation head is designed to replace the by-default flat classification head in Equation~\ref{eqn:flat}. It predicts the following four items:
\vspace{-1mm}
\begin{align}
	\boldsymbol{r}_{ij}^{\text{sc}} &= \operatorname{SoftMax}\{\boldsymbol{X}_{ij}^{\top} \boldsymbol{W}^{\text{sc}}\}\label{eqn:sc} \\
	\boldsymbol{r}_{ij}^{\text{geo}} &= \operatorname{SoftMax} \{\boldsymbol{X}_{ij}^{\top} \boldsymbol{W}^{\text{geo}}\}  \cdot \boldsymbol{r}_{ij}^{\text{sc}}[0] \\
	\boldsymbol{r}_{ij}^{\text{pos}} &= \operatorname{SoftMax}\{\boldsymbol{X}_{ij}^{\top} \boldsymbol{W}^{\text{pos}}\} \cdot \boldsymbol{r}_{ij}^{\text{sc}}[1] \\
	\boldsymbol{r}_{ij}^{\text{sem}} &= \operatorname{SoftMax}\{\boldsymbol{X}_{ij}^{\top} \boldsymbol{W}^{\text{sem}}\} \cdot \boldsymbol{r}_{ij}^{\text{sc}}[2] \label{eqn:post}
\end{align}
\vspace{-1mm}
where $\cdot$ represents the scalar product, $\boldsymbol{r}_{ij}^{\text{sc}}\in\mathbb{R}^{4}$ represents the probabilities of the three relation super-categories plus the background class, and $\{ \boldsymbol{r}_{ij}^{c} \mid c \in [\text{geo}, \text{pos}, \text{sem}] \}$ are the joint probabilities under each super-category. Same for $\mathbf{r}_{ji}$ in the other direction.

To train the hierarchical relation head, we apply cross-entropy losses to all four terms in Equation~\ref{eqn:sc}-\ref{eqn:post}. In addition, we use a supervised contrastive loss~\cite{khosla2020supervised} to minimize the distances in embedding space within the same relation class (set $P(ij)$) and maximize those from different classes (set $N(ij)$). Except for those in DETR~\cite{carion2020end}, the total loss $\mathcal{L}$ can be decomposed as the weighted sum of the following terms:
\begin{align}
& \mathcal{L}_{\text{sup\_rel}} = \operatorname{NLL\ Loss}\ \{\boldsymbol{r}_{ij}^{\text{sc}}, \underline{r}_{ij}^{\text{sc}}\} \label{eqn:loss_sc}\\
& \mathcal{L}_{\text{sub\_rel}} = \hspace{-6mm} \sum_{c\in\{\text{geo}, \text{pos}, \text{sem}\}} \hspace{-6mm} \mathbf{1}_{\text{groundtruth is } c} \cdot \operatorname{NLL\ Loss}\ \{\boldsymbol{r}_{ij}^{c}, \underline{r}_{ij}\} \label{eqn:nll} \\ \vspace{-30mm}
& \mathcal{L}_{\text{contrastive}} = \hspace{-2mm} \sum_{p \in P(ij)}\hspace{-2mm} \log \frac{\exp \left(\boldsymbol{X}_{ij}^{\top} \boldsymbol{X}_{p} / \tau \right)}{\sum_{n \in N(ij)} \exp \left(\boldsymbol{X}_{ij}^{\top} \boldsymbol{X}_{n} / \tau \right)} \label{eqn:supcon}
\end{align}
where the underlined quantities denote ground truth values from the dataset, $\mathcal{L}_{\text{sup\_rel}}$ is the loss for relation super-categories, $\mathcal{L}_{\text{sub\_rel}}$ is the loss for detailed relationships, $\mathcal{L}_{\text{contrastive}}$ is the contrastive loss, and $\tau$ is the temperature. $\mathbf{1}$ is an indicator function, meaning that we back-propagate the losses only to the relations under target super-categories. 

The model yields three predicates for each edge, one from each \textit{disjoint} super-category, maintaining the exclusivity among relations within the same super-category. 
All three predicates from each edge will participate in the confidence ranking. Because there will be three times more candidates, we are not trivially relieving the graph constraints~\cite{newell2017pixels, zellers2018neural} to make the task simpler.

It is often the case that two predicates from disjoint super-categories of the same edge will appear within the top $k$ predictions, providing different interpretations of the edge. This design leverages the super-category probabilities to guide the network's attention toward the appropriate conditional output heads, enhancing the interpretability and performance of the system.

\subsection{Commonsense validation}
\begin{figure*}[t]
  \centering
  \includegraphics[width=0.97\linewidth]{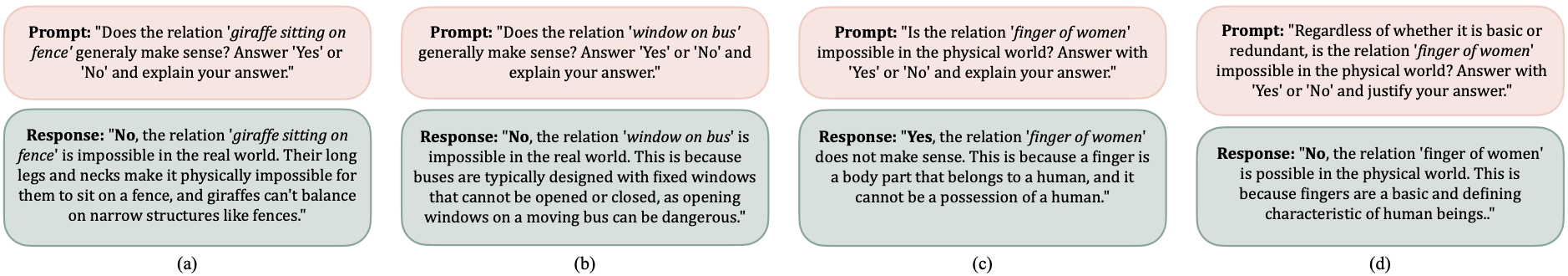}
  \vspace{-2mm}
\caption{Examples of the LLM responses when we query it to validate whether each predicate is reasonable or not and give explanations. (a) shows a success case. (b) shows a failure but the explanation shows that LLMs sometimes tend to consider irrelevant and misleading contexts. (c) and (d) shows how asking for explanations could help design more robust prompts.}
\vspace{-1mm}
  \label{fig:response}
\end{figure*}

Language or vision-language models can critique output predictions of a scene graph generation algorithm and select those that align with common sense. In our settings, we intend to only leverage open-sourced, small-scale language models like LLaMA-3-8B~\cite{touvron2023llama} or small vision-language models like LLaVA-1.6-7B~\cite{liu2023improved} to handle this complex graphical task. This choice is motivated by the potential interests in applying these algorithms on local devices, particularly within the robotics community~\cite{neau2023defense, amodeo2022og, amiri2022reasoning, li2022embodied}, where computations and space are crucial considerations. We also provide a comparison with larger, commercial alternatives like GPT-3.5~\cite{brown2020language} in Section~\ref{sec:experiment} with examples in Figure~\ref{fig:response}.

\begin{figure}[h]
  \centering
  \includegraphics[width=1\linewidth]{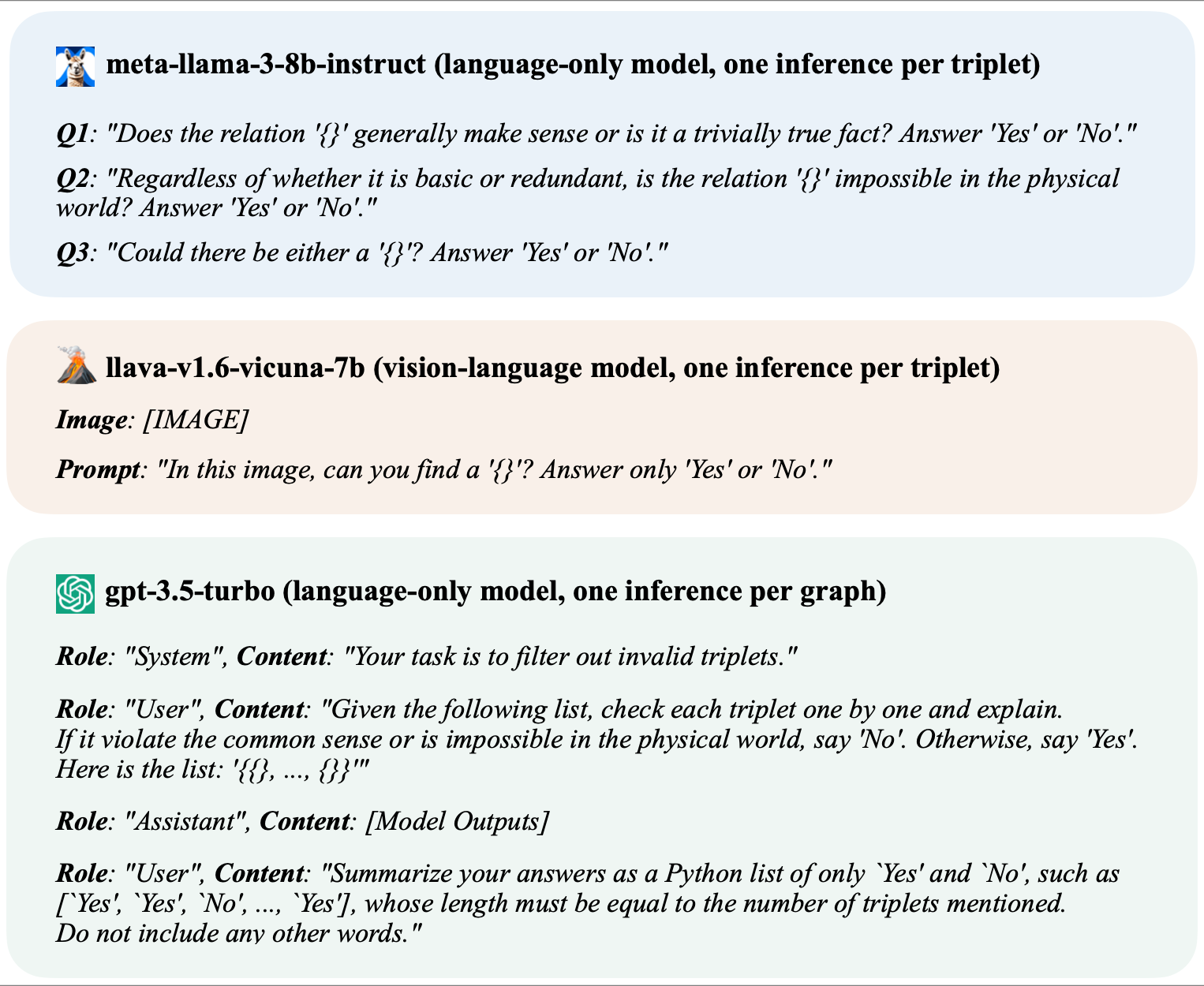}
  \caption{Prompt engineering across different foundation models, where ``\{\}" is a placeholder for a triplet written in string, such as \textit{``girl riding skateboard"}. For LLaMA-3-8B, which lacks the vision capability, we employ three distinct prompts for each of the top $m$-$n$ predicted triplets, and collect a majority vote on whether each triplet makes sense to enhance the robustness. In contrast, LLaVA-1.6-7B is prompted to verify whether each of the top $m$-$n$  predicted triplets actually appears in the image. Since GPT-3.5 is much larger than LLaMA-3-8B with better instruction-following capabilities, we use a single prompt to evaluate all the top $m$-$n$ predicates in the (sub)graph of each image, and collect a list of `Yes' or `No' responses simultaneously with a higer efficiency. }
  \vspace{-3mm}
  \label{fig:prompt}
\end{figure}

By deploying small, open-sourced models, we harness their abilities as a commonsense validator, rather than asking them to generate a scene graph from scratch, which could be more challenging. We query such a foundation model on whether each of the top $m$-$n$ most confident predicates identified per image is reasonable or not using the prompts in Figure~\ref{fig:prompt}. This validation process aims to filter out predictions with high confidence but violate the basic commonsense in the physical world. 

\begin{figure*}[t]
  \centering
  \includegraphics[width=0.96\linewidth]{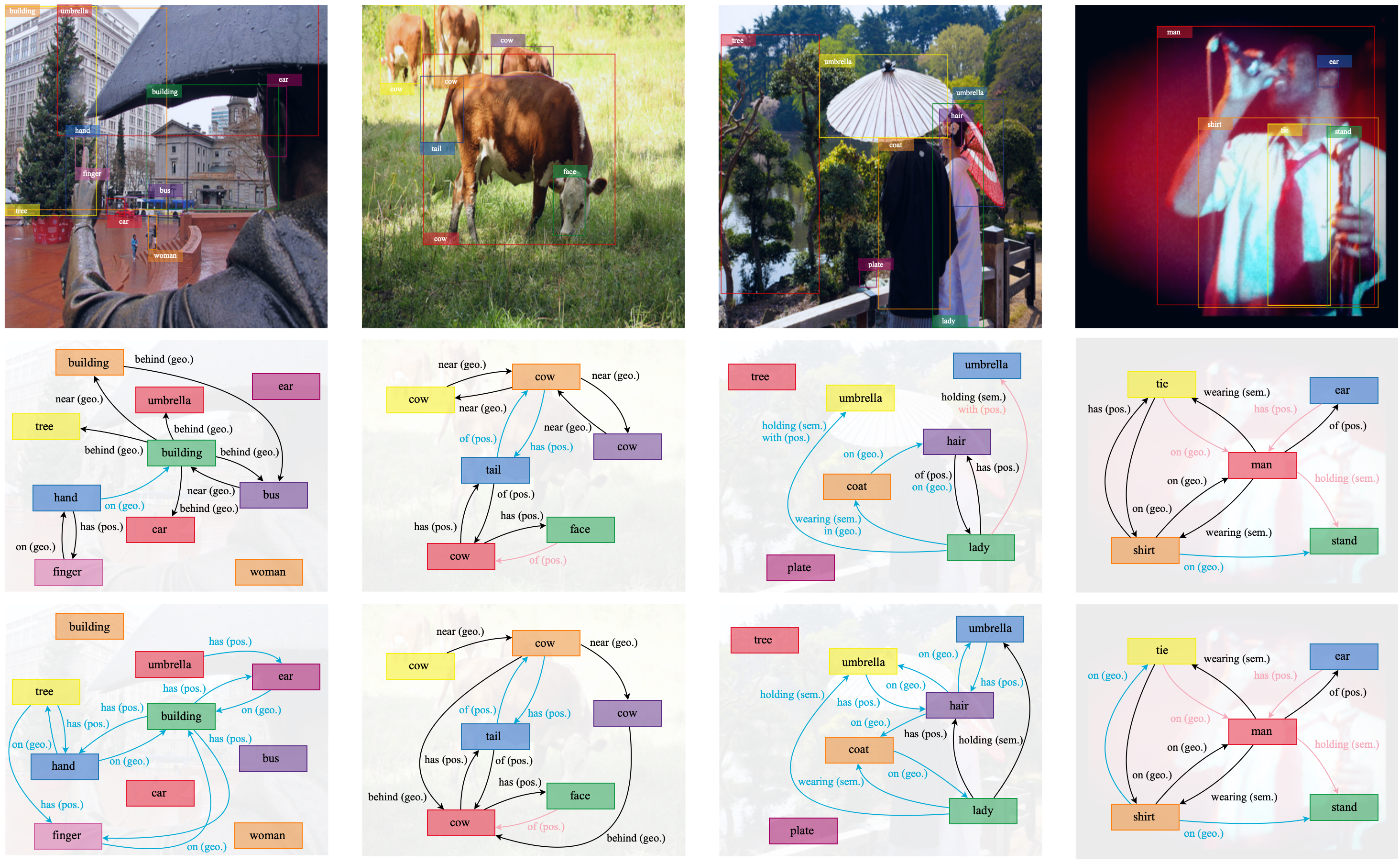}
  \caption{Illustration of generated scene graphs on predicate classification. All examples are from the testing dataset of Visual Genome. The first row displays images and objects, while the second row displays the final scene graphs. The third row shows an ablation without commonsense validation. For each image, we display the top 10 most confident predictions, and each edge is annotated with its relation label and super-category. Meanwhile, it is possible for an edge to have multiple predicted relationships, but they must come from disjoint super-categories. In this figure, \textcolor{mypink}{pink} edges are true positives in the dataset. \textcolor{mydarkblue}{Blue} edges represent incorrect edges based on our observations Interestingly, all the \textbf{black} edges are reasonable predictions we believe but not annotated, which should not be regarded as false positives.}
  \label{fig:plot}
  \vspace{-2mm}
\end{figure*}

\subsection{Seamless integration with existing frameworks}
Figure~\ref{fig:flowchart} illustrates how the proposed HIERCOM - with (1) the ``HIER" part: hierarchical relation head and (2) the ``COM" part: commonsense validation pipeline - can be easily integrated as plug-and-play modules into not only our baseline model but also other existing SOTA algorithms. Specifically, the hierarchical relation head can replace the final linear layer in the classification heads, refining the process of relationship classification. Subsequently, the predicted triplets can undergo the commonsense validation pipeline, eliminating nonsensical ones in the final outputs. 
\section{Experiments}
\label{sec:experiment}

\begin{table*}[h]
  \centering
  \scriptsize
  \caption{This table showcases the main experimental results, presenting Recall@$k$ and mean Recall@$k$ scores from the testing dataset of Visual Genome. We integrate the proposed hierarchical relation head and commonsense validation pipeline into existing scene graph generation algorithms, comparing the performance both with and without these plug-and-play modules. Improvements are evident, with higher scores highlighted in bold, demonstrating the effectiveness of our approaches across a variety of baseline models. For these evaluations, the commonsense validation employs LLaVA-1.6-7B, while results utilizing other foundation models are detailed in Table~\ref{tab:llm_ablation}.}
  \vspace{-4mm}
  \begin{tabular}{c ccc ccc ccc}
    \\ \toprule
    \multicolumn{1}{p{2.8cm}}{} &
    \multicolumn{3}{p{4cm}}{\centering{PredCLS}} &
    \multicolumn{3}{p{4cm}}{\centering{SGCLS}} &
    \multicolumn{3}{p{4cm}}{\centering{SGDET}} \\
    \cmidrule(r){2-4}
    \cmidrule(r){5-7}
    \cmidrule(r){8-10}
    Methods & R@20 & R@50 & R@100 & R@20 & R@50 & R@100 & R@20 & R@50 & R@100 \\
    \midrule
    IMP~\cite{xu2017scene} & - & 44.8 & 53.1 & - & 21.7 & 24.4 & - & 3.4 & 4.2 \\
    HC-Net~\cite{ren2020scene} & 59.6 & 66.4 & 68.8 & 34.2 & 36.6 & 37.3 & 22.6 & 28.0 & 31.2 \\
    GPS-Net~\cite{lin2020gps} & 60.7 & 66.9 & 68.8 & 36.1 & 39.2 & 40.1 & 22.6 & 28.4 & 31.7 \\
    BGT-Net~\cite{dhingra2021bgt} & 60.9 & 67.3 & 68.9 & 38.0 & 40.9 & 43.2 & 23.1 & 28.6 & 32.2 \\
    RelTR~\cite{cong2022reltr} & 63.1 & 64.2 & - & 29.0 & 36.6 & - & 21.2 & 27.5 & - \\
    \midrule
    Baseline (ours) & 59.4 & 67.9 & 69.9 & 29.5 & 33.8 & 34.8 & 20.3 & 26.1 & 28.1 \\
    Baseline+HIERCOM (ours) & \textbf{64.2} & \textbf{75.6} & \textbf{79.2} & \textbf{32.5} & \textbf{37.5} & \textbf{39.2} & \textbf{23.8} & \textbf{29.8} & \textbf{32.7} \\
    \midrule
    Motifs~\cite{zellers2018neural} & \textbf{58.5} & 65.2 & 67.1 & 32.9 & 35.8 & 36.5 & \textbf{21.4} & 27.2 & 30.3 \\
    Motifs+HIERCOM (ours) & 55.5 & \textbf{69.5} & \textbf{75.6} & \textbf{34.0} & \textbf{41.6} & \textbf{44.8} & 20.3 & \textbf{28.3} & \textbf{33.9} \\
    \midrule
    Transformer~\cite{tang2020unbiased} & \textbf{59.0} & 65.7 & 67.6 & \textbf{35.4} & 38.6 & 39.4 & 23.0 & 31.3 & 35.5 \\
    Transformer+HIERCOM (ours) & 54.8 & \textbf{68.5} & \textbf{75.0} & 34.7 & \textbf{41.1} & \textbf{44.3} & \textbf{23.2} & \textbf{31.6} & \textbf{35.9} \\
    \midrule
    VCTree~\cite{tang2019learning} & \textbf{59.8} & 65.9 & 67.6 & \textbf{41.5} & 45.2 & 46.1 & \textbf{24.9} & 32.0 & 36.3 \\
    VCTree+EBM~\cite{suhail2021energy} & 57.3 & 64.0 & 65.8 & 40.3 & 44.7 & 45.8 & 24.2 & 31.4 & 35.9 \\
    VCTree+HIERCOM (ours) & 55.9 & \textbf{69.8} & \textbf{75.8} & 39.0 & \textbf{46.7} & \textbf{50.7} & 23.3 & \textbf{32.2} & \textbf{37.0} \\
    \midrule
    VCTree+TDE~\cite{tang2020unbiased} & 36.2 & 47.2 & 51.6 & 19.9 & 25.4 & 27.9 & 14.0 & 19.4 & 23.2 \\
    VCTree+TDE+HIERCOM (ours) & \textbf{41.1} & \textbf{58.3} & \textbf{67.5} & \textbf{26.3} & \textbf{35.6} & \textbf{40.4} & \textbf{14.5} & \textbf{20.6} & \textbf{25.8} \\
    \midrule
    & mR@20 & mR@50 & mR@100 & mR@20 & mR@50 & mR@100 & mR@20 & mR@50 & mR@100 \\
    \midrule
    IMP~\cite{xu2017scene} & 11.7 & 14.8 & 16.1 & 6.7 & 8.3 & 8.8 & 4.9 & 6.8 & 7.9 \\
    GPS-Net~\cite{lin2020gps} & 17.4 & 21.3 & 22.8 & 10.0 & 11.8 & 12.6 & 6.9 & 8.7 & 9.8 \\
    BGT-Net~\cite{dhingra2021bgt} & 16.8 & 20.6 & 23.0 & 10.4 & 12.8 & 13.6 & 5.7 & 7.8 & 9.3 \\
    RelTR~\cite{cong2022reltr} & 20.0 & 21.2 & - & 7.7 & 11.4 & - & 6.8 & 10.8 & - \\
    \midrule
    Baseline (ours) & 12.1 & 15.1 & 15.8 & 5.8 & 7.2 & 7.8 & 3.2 & 4.6 & 5.5 \\
    Baseline+HIERCOM (ours) & \textbf{17.7} & \textbf{23.9} & \textbf{26.7} & \textbf{9.1} & \textbf{11.7} & \textbf{12.9} & \textbf{4.9} & \textbf{8.2} & \textbf{10.0} \\
    \midrule
    Motifs~\cite{zellers2018neural} & 11.7 & 14.8 & 16.1 & 6.7 & 8.3 & 8.8 & 4.9 & 6.8 & 7.9 \\
    Motifs+HIERCOM (ours) & \textbf{16.3} & \textbf{25.1} & \textbf{30.6} & \textbf{9.8} & \textbf{14.6} & \textbf{17.3} & \textbf{5.7} & \textbf{8.5} & \textbf{10.6} \\
    \midrule
    Transformer~\cite{tang2020unbiased} & 11.6 & 14.7 & 15.8 & 6.7 & 8.2 & 8.7 & 3.7 & 5.0 & 6.0 \\
    Transformer+HIERCOM (ours) & \textbf{17.9} & \textbf{26.6} & \textbf{32.2} & \textbf{10.3} & \textbf{15.1} & \textbf{17.8} & \textbf{6.7} & \textbf{9.6} & \textbf{12.0} \\
    \midrule
    VCTree~\cite{tang2019learning} & 13.1 & 16.5 & 17.8 & 8.5 & 10.5 & 11.2 & 5.3 & 7.2 & 8.4 \\
    VCTree+EBM~\cite{suhail2021energy} & 14.2 & 18.2 & 19.7 & 10.0 & 12.5 & 13.5 & 5.7 & 7.7 & 9.1 \\
    VCTree+HIERCOM (ours) & \textbf{17.6} & \textbf{26.3} & \textbf{31.8} & \textbf{11.8} & \textbf{16.9} & \textbf{20.0} & \textbf{5.8} & \textbf{8.7} & \textbf{11.1} \\
    \midrule
    VCTree+TDE~\cite{tang2020unbiased} & 17.3 & 24.6 & 28.0 & 9.3 & 12.9 & 14.8 & 6.3 & 8.6 & 10.5 \\
    VCTree+TDE+HIERCOM (ours) & \textbf{20.3} & \textbf{29.1} & \textbf{35.5} & \textbf{14.3} & \textbf{20.0} & \textbf{23.7} & \textbf{7.3} & \textbf{10.7} & \textbf{13.7} \\
    \bottomrule
  \end{tabular}
  \label{tab:result_final}
  \vspace{-2mm}
\end{table*}

\subsection{Datasets and evaluation metrics}
Our experiments are conducted on Visual Genome~\cite{krishna2017visual} and OpenImage V6~\cite{kuznetsova2020open} datasets, following the same pre-processing procedures and splits in~\cite{xu2017scene, li2022sgtr}. On Visual Genome, we select the top 150 object labels and 50 relations, resulting in $75.7k$ training and $32.4k$ testing images. We adopt Recall@$k$ (R@$k$) and mean Recall@$k$ (mR@$k$)~\cite{lu2016visual, tang2019learning}. R@$k$ measures the recall within the top $k$ most confident predicates per image, while mR@$k$ computes the average across all relation classes. Zero-shot recall (zsR@$k$)~\cite{tang2020unbiased} calculates R@$k$ for the triplets that only appear in the testing dataset. We conduct three tasks: (1) Predicate classification (PredCLS) predicts relations with ground-truth bounding boxes and labels. (2) Scene graph classification (SGCLS) only assumes known bounding boxes. (3) Scene graph detection (SGDET) has no prior knowledge of objects, while predicted and target boxes should have an IOU of at least 0.5~\cite{lu2016visual}.

On OpenImage V6, we keep $601$ object labels and $30$ relations, resulting in around $53.9k$ training and $3.2k$ testing images. We adopt its standard metrics: Recall@$50$, the weighted mean average precision of relationships $\mathrm{wmAP_{rel}}$ and phrases $\mathrm{wmAP_{phr}}$, and the final score 
    $= 0.2 \cdot R@50 + 0.4 \cdot \mathrm{wmAP_{rel}} + 0.4 \cdot \mathrm{wmAP_{phr}}$, as detailed in~\cite{zhang2019graphical}.

\subsection{Numerical results}
Table~\ref{tab:result_final} highlights the main experimental results. The proposed hierarchical relation head and commonsense validation pipeline are model-agnostic, so we compare the performance on multiple backbone models with and without integrating the proposed modules. Our comparative analysis demonstrates that such a simple incorporation, in almost all cases, enhance the performance by a large margin across all three tasks on both R@$k$~\cite{lu2016visual} and mR@$k$~\cite{tang2019learning} scores, affirming the promising of our approach. Zero-shot results in Table~\ref{tab:zero_shot} also support the same conclusion.

\subsection{Visual results}
Figure~\ref{fig:plot} showcases some predicted scene graphs, which include their top $10$ most confident predicates. The second row displays final scene graphs and the third row shows an ablation without commonsense validation. 
The commonsense validation successfully efficiently reduces many unreasonable predicates that would otherwise have appeared on top of the confidence ranking, for example, it removes the predicate \textit{``tree has hand"} in the left-most image from its initial prediction. 
Furthermore, it is important to note that there are numerous valid predictions aligned with intuitions that are not annotated; these are marked in \textbf{black}, whose number significantly exceeds the number of \textcolor{mypink}{pink} edges representing true positives according to dataset annotations. This discrepancy highlights the sparse nature of the dataset's annotations. Besides, all \textcolor{mydarkblue}{blue} edges are incorrect predictions based on both the dataset annotations and our observations.
Still, our model learns rich relational information even being trained from incomplete annotations. We strongly believe that creating an extensive set of predicates is beneficial for practical scene understanding.

\subsection{More results on the hierarchical relation head}
\paragraph{Training details}
The baseline model is trained using a batch size of 12 for three epochs on Visual Genome or one epoch on OpenImage V6, using two NVIDIA V100 GPUs with 32GB memory. We use an SGD optimizer with a learning rate of 1e-5 and a step scheduler that decreases $lr$ to 1e-6 at the third epoch. We take the DETR backbone from \cite{li2022sgtr} pretrained on the same training dataset and freeze it in our experiments. The model has a size of roughly 1055MB and takes an average of 0.2 seconds on single-image inference. Our plug-and-play experiments are build upon the code framework in~\cite{tang2020unbiased, tang2020sggcode}. We keep training parameters the same, except for learning rates, and retrain all models from scratch with the hierarchical relation head replacing the final linear layer of the original classification head.
\vspace{-1mm}

\begin{figure*}[t]
  \centering
  \includegraphics[width=0.95\linewidth]{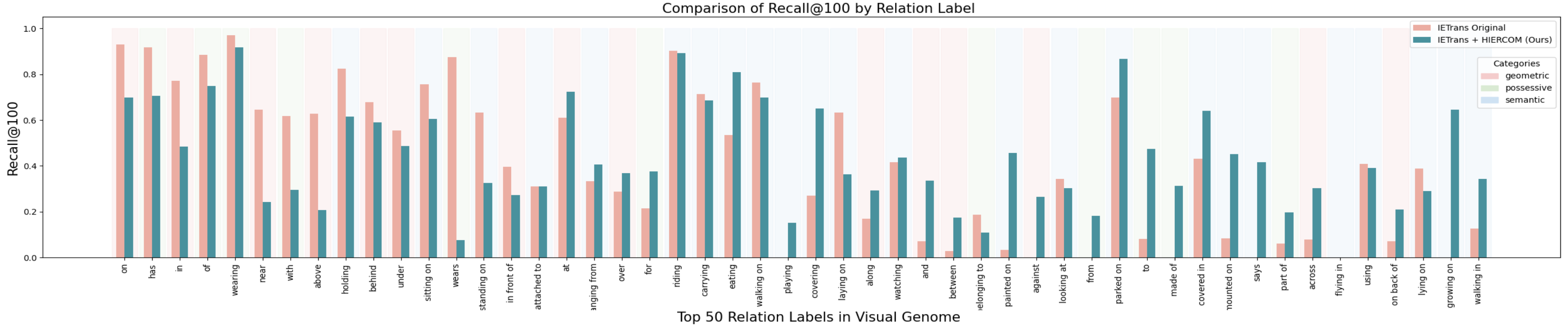}
  \vspace{-2mm}
  \caption{This figure compares the histograms between the original IETrans~\cite{zhang2022fine} and IETrans integrated with the proposed hierarchical relation head and commonsense validation, denoted as IETrans+HIERCOM. Different background colors for each relation label correspond to their super-category, as defined in \cite{zellers2018neural}. While there is a slight decrease in performance for the head classes, continuing improvements are observed in the tail classes. Along with the data in Table~\ref{tab:long_tail}, we show that our proposed methods not only elevate the mR@$k$ scores but also maintain a good balance between the head and tail classes, simultaneously enhancing the R@$k$ scores.}
  \label{fig:histogram}
  \vspace{-3mm}
\end{figure*}

\paragraph{Handling the long-tailed distribution}\label{sec:long_tail}
We integrate the proposed modules into algorithms that are specifically tailored to address the long-tailed distribution of relation labels~\cite{li2022devil, zhang2022fine}. We also present the performance of CogTree~\cite{yu2020cogtree} and HML~\cite{deng2022hierarchical}. Since both methods employ their own hierarchical structures, we did not integrate our hierarchical prediction head to avoid unnecessary complexity. While their methods achieve higher mR@$k$ scores, they exhibit significantly lower R@$k$ scores. Numerical results in Table~\ref{tab:long_tail} shows that our model-agnostic modules can continue raising their SOTA mR@$k$ scores while \textit{simultaneously} achieving even higher R@k scores. Furthermore, the histograms in Figure~\ref{fig:histogram} illustrate that we can gain further improvements on the labels at the tail of the distribution.
\vspace{-1mm}

\begin{table}[t]
  \centering
  \scriptsize
  \caption{This table focuses on methods tailored to tackle the long-tailed distribution problem of relation labels. We show results from the PredCLS task. \cite{yu2020cogtree, deng2022hierarchical, yu2023visually} employ their own hierarchical structures, so we did not integrate our hierarchical prediction head to avoid unnecessary redundancy. \cite{yu2020cogtree, deng2022hierarchical} achieve a different balance with much lower R@$k$ but higher mR@$k$ than HIERCOM on Motifs, Transformer, and VCTree in Table~\ref{tab:result_final}. Besides, algorithms addressing long-tailed distributions typically compromise R@$k$ scores significantly; however, results on~\cite{li2022devil, zhang2022fine} demonstrate that incorporating the relation hierarchy not only continues to improve mR@$k$ scores but also enhances their R@$k$ simultaneously.}
  \vspace{-3mm}
  \begin{tabularx}{\columnwidth}{cYYYY}
    \\ \toprule
    Methods\hspace{-3mm} & R@50 & R@100 & mR@50 & mR@100 \\
    \midrule
    Motifs+CogTree~\cite{yu2020cogtree} & 35.6 & 36.8 & 26.4 & 29.0 \\
    Transformer+CogTree~\cite{yu2020cogtree} & 38.4 & 39.7 & \textbf{28.4} & \textbf{31.0} \\
    VCTree+CogTree~\cite{yu2020cogtree} & \textbf{44.0} & \textbf{45.4 }& 27.6 & 29.7 \\
    \midrule
    Motifs+HML~\cite{deng2022hierarchical} & \textbf{47.1} & \textbf{49.1} & 36.3 & 38.7 \\
    Transformer+HML~\cite{deng2022hierarchical} & 45.6 & 47.8 & 33.3 & 35.9 \\
    VCTree+HML~\cite{deng2022hierarchical} & 47.0 & 48.8 & \textbf{36.9} & \textbf{39.2} \\
    \midrule
    Motifs+CaCao~\cite{yu2023visually} & - & - & 37.1 & 38.9 \\
    Motifs+CaCao+HIERCOM (ours) & - & - & \textbf{37.6} & \textbf{39.9} \\
    \midrule
    Motifs+NICE~\cite{li2022devil} & 55.1 & 57.2  & 29.9 & 32.3 \\
    Motifs+NICE+HIERCOM (ours) & \textbf{58.2} & \textbf{65.4} & \textbf{33.1} & \textbf{39.8} \\
    \midrule
    Motifs+IETrans~\cite{zhang2022fine} & 48.6 & 50.5 & 35.8 & 39.1 \\
    Motifs+IETrans+HIERCOM (ours) & \textbf{60.4} & \textbf{66.4} & \textbf{38.0} & \textbf{44.1} \\
    \bottomrule
  \end{tabularx}
  \label{tab:long_tail}
\end{table}

\paragraph{Automatic clustering of the relation hierarchy}
We discuss an alternative design choice that automatically clusters the relation hierarchy. Instead of adhering to a manually defined hierarchy outlined in Neural Motifs~\cite{zellers2018neural}, the clustering procedure does not necessarily require human involvement. For example, we can employ k-means, a classic unsupervised clustering technique, on pretrained embedding space to construct the relation hierarchy autonomously. 
Table~\ref{tab:embedding} shows results using the CLIP-Text~\cite{radford2021learning}, GPT-2~\cite{brown2020language}, and BERT~\cite{devlin2018bert} embedding space by transforming relation labels as words into corresponding token embeddings. The commonsense validation is also seamlessly integrated here. It turns out that CLIP performs the closest to the manual one in~\cite{zellers2018neural} with even higher mR@k scores, so CLIP could be utilized to generalize our idea to new datasets at any scale without the need for manual clustering.
\vspace{-1mm}

\begin{table}[t]
  \centering
  \scriptsize
  \caption{This table presents ablation studies on different clustering methods used to structure the relation hierarchy. Results are evaluated on Baseline+HIERCOME. We compare manual clustering defined in~\cite{zellers2018neural} that categorizes relations into geometric, possessive, and semantic super-categories, with automatic clustering approaches. The latter ones transform relation labels into CLIP-Text, GPT-2, or BERT's token embedding space and apply k-means clustering with $k=3$. Results are comparably strong, suggesting that the proposed hierarchical relation head can be effectively generalized to new, larger datasets using automatic clustering methods, thus eliminating the need for manual effort.}
  \vspace{-3mm}
  \begin{tabularx}{\columnwidth}{cYYYYYY}
    \\ \toprule
    Methods (all ours)\hspace{-3mm} & R@20\hspace{-3mm} & R@50 & R@100 & mR@20 & mR@50 & mR@100 \\
    \midrule
    Manual~\cite{krishna2017visual} & \textbf{64.2} & \textbf{75.6} & \textbf{79.2} & 17.7 & 23.9 & 26.7 \\
    CLIP-Text ~\cite{radford2021learning} & 61.9 & 73.6 & 77.4 & 17.1 & \textbf{25.6} & \textbf{30.5} \\
    GPT-2~\cite{brown2020language} & 62.5 & 70.5 & 72.7 & \textbf{17.9} & 25.0 & 29.7 \\
    BERT~\cite{devlin2018bert} & 61.8 & 69.9 & 72.8 & 16.3 & 23.3 & 27.4 \\
    \bottomrule
  \end{tabularx}
  \label{tab:embedding}
  \vspace{-5mm}
\end{table}

\begin{table}[t]
  \centering
  \scriptsize
  \caption{Additional ablation studies on Baseline+HIER without the commonsense validation pipeline. We present PredCLS results on Visual Genome to examine the impact of removing depth maps and contrastive loss from the training process.}
  \vspace{-4mm}
  \begin{tabularx}{\columnwidth}{cYYYYYY}
    \\ \toprule
    Methods (all ours)\hspace{-3mm} & R@20 & R@50 & R@100 & mR@20 & mR@50 & mR@100 \\
    \midrule
    Baseline+HIER  & \textbf{61.1} & \textbf{73.6} & \textbf{78.1} & \textbf{14.4} & \textbf{20.6} & \textbf{23.7} \\
    -Depth Maps & 60.0 & 72.2 & 76.9 & 13.3 & 20.1 & 23.1 \\
    -$\mathcal{L}_{\text{contrastive}}$ in Eqn~\ref{eqn:supcon} & 59.1 & 72.9 & 77.6 & 11.7 & 17.2 & 20.1 \\
    \bottomrule
  \end{tabularx}
  \label{tab:more_ablation}
  \vspace{-2mm}
\end{table}

\begin{table}[t]
  \centering
  \scriptsize
  \caption{Zero-shot relationship retrieval~\cite{tang2020unbiased, lu2016visual} results on Visual Genome. We evaluate the zsR@$50$/$100$ for each of the three tasks.}
  \vspace{-4mm}
  \begin{tabularx}{\columnwidth}{cYYY}
    \\ \toprule
    Methods & PredCLS & SGCLS & SGDET \\
    \midrule
    Motifs~\cite{zellers2018neural} & 10.9/14.5 & 2.2/3.0 & 0.1/0.2 \\
    Motifs+HIERCOM (ours) & \textbf{18.4}/\textbf{25.5} & \textbf{4.9}/\textbf{6.4} & \textbf{2.1}/\textbf{3.1} \\
    \midrule
    Transformer~\cite{zhang2017visual} & 11.3/14.7 & 2.5/3.3 & 0.8/1.5 \\
    Transformer+HIERCOM (ours) & \textbf{20.1}/\textbf{26.8} & \textbf{5.6}/\textbf{7.3} & \textbf{1.9}/\textbf{3.1} \\
    \midrule
    VCTree~\cite{tang2019learning} & 10.8/14.3 & 1.9/2.6 & 0.2/0.7 \\
    VCTree+HIERCOM (ours) & \textbf{17.8}/\textbf{24.8} & \textbf{6.6}/\textbf{9.4} & \textbf{2.0}/\textbf{3.1} \\
    \midrule
    VCTree+TDE~\cite{tang2020unbiased} & \textbf{14.3}/17.6 & 3.2/4.0 & \textbf{2.6}/\textbf{3.2} \\
    VCTree+TDE+HIERCOM (ours) & 13.7/\textbf{20.2} & \textbf{4.3}/\textbf{6.2} & 1.7/2.5 \\
    \bottomrule
  \end{tabularx}
  \label{tab:zero_shot}
  \vspace{-2mm}
\end{table}

\begin{table}[t]
  \centering
  \scriptsize
  \caption{Predicate classification results on OpenImage V6.}
  \vspace{-5mm}
  \begin{tabularx}{\columnwidth}{cYYYY}
    \\ \toprule
    Methods & R@50 & $\mathrm{wmAP_{rel}}$ & $\mathrm{wmAP_{phr}}$ & $\mathrm{score}$ \\
    \midrule
    SGTR~\cite{carion2020end} & 59.9 & \textbf{37.0} & 38.7 & 42.3 \\
    RelDN~\cite{zhang2019graphical} & 72.8 & 29.9 & 30.4 & 38.7 \\
    GPS-Net~\cite{lin2020gps} & 74.7 & 32.8 & 33.9 & 41.6 \\
    \midrule
    Baseline+HIER (ours) & \textbf{85.4} & 33.1 & \textbf{44.9} & \textbf{48.3} \\
    \bottomrule
  \end{tabularx}
  \vspace{-3mm}
  \label{tab:oiv6}
\end{table}

\paragraph{More ablation studies}
The first row of each section in Table~\ref{tab:llm_ablation} shows the ablation with the relation hierarchy but without the commonsense validation. Table~\ref{tab:more_ablation} details the impact of incorporating depth maps and the supervised contrastive loss in Equation~\ref{eqn:supcon} on our baseline model. 
\vspace{-1mm}

\paragraph{Results on the OpenImage V6 dataset}
Numerical results from the OpenImage V6 dataset again validate the effectiveness of our proposed hierarchical relation head, as shown in Table~\ref{tab:oiv6}. It underscores the consistent performance enhancements achieved across different datasets.


\begin{table*}[h]
  \centering
  \scriptsize
  \smaller
  \caption{This table provides a comprehensive ablation study on (1) models with hierarchical relation head but without commonsense validation, i.e., ``+HIER" in the first row of each section, and (2) the choice of foundation models for commonsense validatio, including LLaMA-3-8B~\cite{touvron2023llama}, GPT-3.5~\cite{brown2020language}, and LLaVA-1.6-7B~\cite{liu2023improved}. The performances across these models are largely comparable, with LLaVA-1.6-7B showing a slight advantage. This finding is particularly promising as it suggests that our commonsense validation pipeline can be effectively deployed using only a small-scale, open-sourced language model on local devices, making it accessible for practical use.}
  \vspace{-3mm}
  \begin{tabular}{c ccc ccc ccc}
    \\ \toprule
    \multicolumn{1}{p{2cm}}{} &
    \multicolumn{3}{p{3cm}}{\centering{PredCLS}} &
    \multicolumn{3}{p{3cm}}{\centering{SGCLS}} &
    \multicolumn{3}{p{3cm}}{\centering{SGDET}} \\
    \cmidrule(r){2-4}
    \cmidrule(r){5-7}
    \cmidrule(r){8-10}
    Methods (all ours) & R@20 & R@50 & R@100 & R@20 & R@50 & R@100 & R@20 & R@50 & R@100 \\
    \midrule
    Motifs+HIER 
    & 53.8 & 68.3 & 74.6 & 30.6 & 36.0 & 37.6 & \textbf{22.8} & \textbf{29.3} & 32.1 \\
    +LLaMA/GPT/LLaVA
    & 55.2/55.4/\textbf{55.5} & 69.5/\textbf{69.6}/69.5 & 75.7/75.7/\textbf{75.6} & 33.8/33.9/\textbf{34.0} & 41.5/\textbf{41.6}/\textbf{41.6} & \textbf{44.8}/\textbf{44.8}/\textbf{44.8} & 20.0/20.1/20.3 & 27.9/28.2/28.3 & 33.5/\textbf{33.9}/\textbf{33.9} \\
    \midrule
    Transformer+HIER & 53.8 & 68.1 & 74.5 & 33.9 & \textbf{41.4} & \textbf{44.7} & 20.0 & 28.1 & 34.9 \\
    +LLaMA/GPT/LLaVA
    & 53.5/53.5/\textbf{54.8} & \textbf{68.5}/68.4/\textbf{68.5} & \textbf{75.1}/75.0/75.0 & 33.5/33.6/\textbf{34.7} & 41.1/41.1/41.1 & 44.4/44.3/44.3 & 20.1/19.6/\textbf{23.2} & 28.1/29.8/\textbf{31.6} & 35.0/35.7/\textbf{35.9} \\
    \midrule
    VCTree+HIER & 54.5 & 69.1 & 75.4 & 38.0 & 46.4 & 50.5 & 19.5 & 26.9 & 32.6 \\
    +LLaMA/GPT/LLaVA
    & 55.5/55.5/\textbf{55.9} & \textbf{69.8}/\textbf{69.8}/\textbf{69.8} & 75.8/\textbf{76.0}/75.8 & 37.8/38.1/\textbf{39.0} & 46.6/\textbf{46.7}/\textbf{46.7} & 50.7/\textbf{50.8}/50.7 & 21.0/21.1/\textbf{23.3} & 30.0/30.2/\textbf{32.2} & 35.9/36.1/\textbf{37.0} \\
    \midrule
    & mR@20 & mR@50 & mR@100 & mR@20 & mR@50 & mR@100 & mR@20 & mR@50 & mR@100 \\
    \midrule
    Motifs+HIER 
    & 15.9 & 24.3 & 29.9 & 7.7 & 10.4 & 11.9 & 4.1 & 6.8 & 8.7 \\
    +LLaMA/GPT/LLaVA
    & 16.1/16.2/\textbf{16.3} & 25.0/\textbf{25.1}/\textbf{25.1} & 30.6/\textbf{30.8}/30.6 & 9.7/\textbf{9.8}/\textbf{9.8} & \textbf{14.6}/14.5/\textbf{14.6} & 17.3/\textbf{17.4}/17.3 & \textbf{5.7}/5.6/\textbf{5.7} & 8.4/\textbf{8.5}/\textbf{8.5} & 10.6/\textbf{10.7}/10.6 \\
    \midrule
    Transformer+HIER & \textbf{18.1} & 26.2 & 31.5 & \textbf{10.4} & 15.0 & 17.7 & 6.6 & 9.6 & 12.0 \\
    +LLaMA/GPT/LLaVA
    & 17.7/17.8/17.9 & 26.4/\textbf{26.7}/26.6 & 32.2/\textbf{32.4}/32.2 & 10.2/10.3/10.3 & \textbf{15.1}/\textbf{15.1}/\textbf{15.1} & 17.8/\textbf{17.9}/17.8 & 6.6/\textbf{6.7}/\textbf{6.7} & 9.6/\textbf{9.7}/9.6 & 12.0/\textbf{12.1}/12.0 \\
    \midrule
    VCTree+HIER & 16.7 & 26.1 & \textbf{32.2} & 11.5 & 16.8 & 20.0 & 5.7 & 8.5 & 10.7 \\
    +LLaMA/GPT/LLaVA
    & 17.3/17.4/\textbf{17.6} & \textbf{26.3}/\textbf{26.3}/\textbf{26.3} & 31.8/31.9/31.8 & 11.7/\textbf{11.8}/\textbf{11.8} & 16.8/\textbf{16.9}/\textbf{16.9} & 20.1/\textbf{20.2}/\textbf{20.2} & 5.5/5.7/\textbf{5.8} & 8.6/8.6/\textbf{8.7} & 11.0/\textbf{11.1}/\textbf{11.1} \\
    \bottomrule
  \end{tabular}
  \label{tab:llm_ablation}
  \vspace{-2mm}
\end{table*}

\subsection{More results on the commonsense validation}

\paragraph{Inference details}
We implement the commonsense validation pipeline on LLaMA-3-8B~\cite{touvron2023llama} and LLaVA-1.6-7B~\cite{liu2023improved} on a single NVIDIA V100 GPU with 32GB memory (around 3.37s and 4.75s per image on Motifs+HIERCOM), and GPT-3.5~\cite{brown2020language} through its API calls. 
We skip the top $m=10$ most confident predictions and validate the next $20$ ones per image. To minimize LLM's hallucinations~\cite{huang2023survey}, we exclude validating any subject-relation-object combinations in the training dataset, resulting in $0\%$ unique false-negative triplets in training and $0.27\%$ in testing.

\paragraph{Different choices of foundation models}
Table~\ref{tab:llm_ablation} presents comprehensive ablation studies on utilizing LLMs at different scales and a VLM with vision capabilities. Surprisingly, the results indicate only minor differences in the performance among these models on filtering out commonsense-violated predictions. Such an uniform effectiveness and robustness suggest that users can effectively implement our commonsense validation pipeline using even a small-scale, open-sourced, and language-only model like LLaMA-3-8B~\cite{touvron2023llama} on their local devices, ensuring a good accessibility.

\paragraph{Commonsense distillation}
To address slowdowns in inference caused by incorporating LLMs, we explore the option of embedding commonsense knowledge from LLMs directly into our baseline model. During an initial training phase, we categorize predicates into those that align with commonsense ($\mathcal{S}_\text{aligned}$) and those that violate it ($\mathcal{S}_\text{violated}$). After this classification, we can retrain the same model from scratch but with an additional loss
$\mathcal{L}_{\mathrm{cs}}=\mathbf{1}_{\notin \mathcal{S}_{\text {aligned }}} * \lambda_{\text {weak }}+\mathbf{1}_{\in \mathcal{S}_{\text {violated }}} * \lambda_{\text {strong }}$.
Here, $\lambda_{\text {weak }}$=$0.1$ and $\lambda_{\text {strong }}$=$10$ serve as penalty weights, and $\mathbf{1}$ is an indicator. 
We find only little difference within $1\%$ between querying LLMs at inference and using distilled models. Alternative to distillation, we can also filter out predicates based on $\mathcal{S}_\text{violated}$ during future runs of inference. Both options are encouraging since they eliminates the necessity to access LLMs at test time.

\section{Conclusion}
\label{sec:conclusion}
In this study, we demonstrate that leveraging hierarchical relationships significantly enhances performance in scene graph generation. Additionally, our proposed commonsense validation pipeline effectively filters out predicates that violate commonsense knowledge, even when implemented with a small-scale, open-sourced language model. Both approaches are model-agnostic and can be seamlessly integrated into a variety of existing scene graph generation models to push their performance to new levels.

\section{Acknowledgment}
This work was supported by the National Science Foundation (NSF) grant CCF-2112665 (TILOS).

{\small
\bibliographystyle{ieee_fullname}
\bibliography{egbib}
}

\end{document}